\documentclass[conference]{IEEEtran}
\IEEEoverridecommandlockouts
\usepackage{cite}
\usepackage{float}
\usepackage{amsmath,amssymb,amsfonts}
\usepackage{algorithmic}
\usepackage{graphicx}
\usepackage{textcomp}
\usepackage{xcolor}
\usepackage{booktabs}
\usepackage{url}
\usepackage{newtxtext}
\def\BibTeX{{\rm B\kern-.05em{\sc i\kern-.025em b}\kern-.08em
    T\kern-.1667em\lower.7ex\hbox{E}\kern-.125emX}}

\newif\ifblind
\blindfalse

\begin{document}

\title{Does Personalization Break Trust? Open-Set Rejection Calibration Under Few-Shot On-Device Intent Adaptation}

\ifblind
\author{\IEEEauthorblockN{Anonymous Author(s)}
\IEEEauthorblockA{Submitted for double-blind review}
}
\else
\author{
\IEEEauthorblockN{Vaibhav Prajapati}
\IEEEauthorblockA{University of Technology Nuremberg\\
Nuremberg, Germany\\
Email: vaibhav.prajapati@utn.de}
\and
\IEEEauthorblockN{Pushp}
\IEEEauthorblockA{Indian Institute of Information Sri City\\
Chittoor, India\\
Email: pushp.g23@iiits.in}
}
\fi

\maketitle

\begin{abstract}
Always-on virtual assistants increasingly personalize on-device to new user-specific commands via few-shot fine-tuning (e.g., LoRA), but this adaptation is evaluated almost exclusively on whether it learns the new commands, not on whether the assistant can still recognize an input that matches \emph{none} of its intents - the out-of-scope (OOS) rejection problem. We show that naive few-shot LoRA personalization degrades OOS-rejection AUROC (area under the ROC curve) on every one of five public intent-classification benchmarks (mean drop $0.055$--$0.258$, five seeds each), while in-scope accuracy is preserved. A scorer-invariant check shows part of this drop is an artifact of the standard confidence scorer's own miscalibration under adaptation, though a smaller, genuine effect remains under any scorer; we trace it to a measurable shrinkage of the embedding margin around known-intent regions (pooled Spearman $\rho=0.61$). A deployment-cheap fix is scoring rejection by Mahalanobis distance instead of softmax confidence without extra training/retraining recovers most of the lost quality on 3 of 5 datasets at no accuracy cost, beats two further OOD scoring baselines, and generalizes across shot count, LoRA rank, and a second base encoder. We report honestly where the fix falls short and release all code for reproducibility.
\end{abstract}

\begin{IEEEkeywords}
pervasive computing, on-device personalization, out-of-scope detection, low-rank adaptation, intent classification, trust calibration
\end{IEEEkeywords}

\section{Introduction}

The oldest promise of ubiquitous computing ambient assistance woven invisibly into daily life \cite{b_weiser} is now largely delivered through voice-driven conversational interfaces: phone assistants, smart speakers, in-car systems, and smart-home hubs that map spoken or typed utterances to a fixed set of known intents. To feel truly personal, these systems increasingly adapt \emph{on-device}, after deployment, to commands specific to one household or one user: a nickname for a routine, a locally-named device, a personal shortcut phrase. Because on-device compute and memory are scarce, this adaptation is typically done with parameter-efficient fine-tuning techniques, most commonly low-rank adaptation (LoRA) \cite{b_lora} by using only a handful of labelled examples per new intent.

\begin{figure}[t]
\centering
\includegraphics[width=\linewidth]{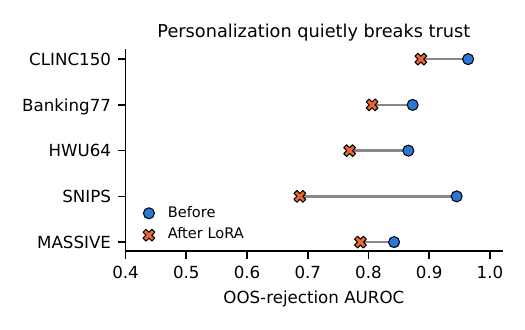}
\caption{Naive few-shot LoRA personalization degrades OOS-rejection AUROC on every one
of 5 intent benchmarks (mean over 5 seeds), while in-scope accuracy is preserved.}
\label{fig:teaser}
\end{figure}

This adaptation step is evaluated, almost universally, on one axis: does the assistant now correctly recognize the new intents it was just taught? A second, equally important axis is almost never checked: does the assistant still correctly recognize when an input belongs to \emph{none} of its intents at all - known or new? This is the out-of-scope (OOS) rejection problem, and it is a trust-critical property of any deployed pervasive assistant. An assistant that silently loses its ability to say ``I don't understand that'' will instead confidently misroute background chatter, a different household member's unrelated request, or a misheard utterance into one of its newly taught commands. Prior work on voice-assistant failures has found that exactly this kind of over-eager acceptance is disproportionately damaging to user trust \cite{b_trust}, and a concurrent 2025 benchmark on smart-home query rejection confirms this remains an unsolved, actively studied problem in deployed pervasive systems \cite{b_reject}.

This paper asks and answers, with a controlled empirical study that \textbf{does few-shot on-device LoRA personalization, as an unintended side effect of teaching new intents, degrade OOS rejection quality?} The closest prior findings establish related but distinct phenomena. Continual-learning work on conformal calibration under repeated LLM fine-tuning shows that coverage guarantees can collapse well before task accuracy does \cite{b_calib}, but conformal coverage is a fundamentally \emph{closed-world} notion (is the true label among the known classes in a prediction set) - it says nothing about whether a genuinely unknown input is correctly flagged as unknown. Work on parameter-efficient on-device personalization \cite{b_prisp} and on-device federated vocabulary personalization \cite{b_nis} both show that personalization introduces new failure surfaces, but neither studies OOS rejection. The orthogonal-projection LoRA family \cite{b_oplora} and the broader open-world continual learning literature \cite{b_owcl} target preserving classification accuracy on old classes under continual adaptation, not preserving calibration against unknown inputs. We bring these threads together in a setting none of them individually covers: text-based, few-shot, LoRA-based on-device personalization, evaluated specifically for OOS-rejection calibration.

We make four contributions, each validated across five public intent-classification benchmarks (three of which - SNIPS \cite{b_snips}, MASSIVE \cite{b_massive}, and the 21-domain HWU corpus \cite{b_hwu} were built directly from or for deployed voice-assistant platforms) and five random seeds per configuration:

\begin{itemize}
\item \textbf{C1 -- Characterizing the drop.} We show naive few-shot LoRA personalization degrades OOS-rejection AUROC on every one of five datasets, with a 95\% confidence interval excluding zero drop in every case; a scorer-invariant check further shows the drop's magnitude is substantially inflated by the standard confidence scorer's own miscalibration under adaptation, though a smaller, same-direction effect remains when the same Mahalanobis-distance  corer is used at both measurement points instead (Section~\ref{sec:results-c1}).

\item \textbf{C2 -- A mechanistic explanation.} We show the drop is associated with a measurable shrinkage in the embedding margin between known-intent class centroids and true out of-scope points, using a Mahalanobis-distance formulation grounded in established feature space out-of-distribution detection theory \cite{b_mahalanobis,b_openmax} Section~\ref{sec:results-c2}).

\item \textbf{C3 -- A validated, low-cost fix.} We test two literature-grounded fix components: an orthogonal-projection constraint on the LoRA update \cite{b_oplora} and a calibration-replay-style threshold recalibration \cite{b_calib} -- and find the latter does not work in this setting. We then show that simply replacing the evaluation-time rejection score (softmax confidence) with a Mahalanobis-distance score recovers most of the lost OOS quality on 3 of 5 datasets, with no in-scope accuracy cost, no additional training, and a quantified low-millisecond on-device latency cost, beating two further established OOD scorers (energy score \cite{b_energy}, ODIN \cite{b_odin}) on every dataset (Section~\ref{sec:results-c3}).

\item \textbf{C4 -- Generality.} We show the effect and the fix are not artifacts of one fixed hyperparameter setting: both hold across few-shot budget ($k\in\{1,5,10\}$), LoRA rank ($r\in\{4,8,16,32\}$), and a second base encoder, on 8 of 9 tested conditions (Section~\ref{sec:results-c4}).
\end{itemize}

Section~\ref{sec:limitations} states plainly where the fix does not fully close the gap, and we release all training, evaluation, and figure-generation code for full reproducibility.\footnote{Anonymized code: \url{https://anonymous.4open.science/r/percom-FE39}}

\section{Related Work}

\textbf{Calibration under continual/incremental adaptation.} The closest sibling finding to
this paper's C1 is Continual Calibration \cite{b_calib}, which shows that conformal coverage
collapses roughly $3.4\times$ faster than task accuracy across repeated LLM fine-tuning
rounds, and proposes ``calibration replay'' -- refitting a conformal threshold from a small
held-out buffer after each update -- as a lightweight fix. Their coverage guarantee is,
however, a closed-world notion: the prediction set is always drawn from the task's known
label space, so their protocol contains no genuinely out-of-scope input at any point. This
paper studies the structurally different question of whether an adapted model can still tell
that an input belongs to \emph{no} known class, and directly adapts their calibration-replay
mechanism as one candidate fix component (Section~\ref{sec:method}), finding that it does not
transfer to the OOS setting.

\textbf{Orthogonal/null-space-constrained LoRA.} An active family of methods constrains LoRA
updates to an orthogonal complement of a frozen weight's dominant subspace to prevent
catastrophic forgetting under continual fine-tuning \cite{b_oplora}. These methods are
evaluated exclusively for classification-accuracy retention on held-out tasks; none evaluates
open-set/OOS-rejection preservation. We adapt the orthogonal-projection mechanism of
\cite{b_oplora} as one component of our fix.

\textbf{On-device / few-shot personalization.} PRISP \cite{b_prisp} personalizes LLMs
on-device via a hypernetwork-generated LoRA under strict compute and privacy budgets, and
\cite{b_nis} personalizes on-device language models' vocabulary via federated learning,
finding that closed-vocabulary on-device models trade coverage of common patterns for an
inability to handle personal, out-of-vocabulary words -- a structurally similar
``personalization fixes one thing, breaks another'' finding to this paper's own, in a
different failure mode. Separately, \cite{b_plmcompress} targets mobile-deployed few-shot
intent detectors from a model-compression angle, optimizing size and in-scope accuracy
without examining OOS behavior after adaptation. None of these three studies OOS-rejection
calibration.

\textbf{Out-of-scope and open-set detection.} Feature-space distance-based OOD detection is
well established: Lee~et~al.\ \cite{b_mahalanobis} fit class-conditional Gaussians on
penultimate-layer features and score inputs by Mahalanobis distance to the nearest class
mean; Bendale and Boult's OpenMax \cite{b_openmax} calibrates a rejection score from
per-class distance-to-mean-activation-vector statistics. We adapt \cite{b_mahalanobis}'s
formulation as our evaluation-time OOS scorer (the fix component that works,
Section~\ref{sec:method}), consistent with contrastive-learning findings that a shrinking
representation margin degrades Mahalanobis-style OOS detection in pretrained transformers
\cite{b_contrastive}. A concurrent vision-domain finding reinforces this paper's core
mechanism from a different angle: Mahalanobis-style OOD-detector reliability shifts
substantially under different fine-tuning regimes \cite{b_geometry}, consistent with this
paper's C2 finding that personalization itself perturbs the feature-space geometry a
Mahalanobis scorer depends on -- though \cite{b_geometry} studies pretrained-backbone
variation in vision models, not a before/after adaptation comparison in the text-intent
setting studied here.

We additionally compare against two further established scorers as baselines: the energy
score \cite{b_energy}, computed directly from logits with no retraining required, and ODIN's
temperature-scaling half \cite{b_odin} (its input-perturbation step is not reproduced here, a
disclosed scoping limit; see Section~\ref{sec:results-c3}). DROID \cite{b_droid} is a strong,
purpose-built dual-encoder OOS detector evaluated on CLINC150 and Banking77 -- two of the
five datasets used here -- but strictly as a static, one-shot detector, never before/after
any adaptation step; its absolute OOS quality is therefore not directly comparable to the
plain-encoder numbers reported in this paper.

\textbf{Rejection in deployed voice assistants.} A concurrent 2025 benchmark studies
personalized query rejection for Chinese smart-home voice assistants using a retrieval- and
history-based method on top of a fixed large-language-model base \cite{b_reject}. This is
the closest work situating OOS rejection explicitly in a pervasive, personalized-assistant
setting; it studies the opposite causal direction from this paper -- personalization
\emph{designed to improve} rejection via conversational history -- rather than rejection
quality degrading as a side effect of ordinary few-shot adaptation, and uses a different
(retrieval-augmented, LLM-based) architecture with no dataset overlap. Together with a 2023
mixed-methods study finding that ``overcapturing'' failures (accepting input that should have
been rejected) disproportionately damage user trust in voice assistants \cite{b_trust}, this
confirms that OOS-rejection reliability under personalization is a live, practically
important problem for pervasive systems, not a synthetic one.

\section{Problem Formulation}
\label{sec:problem}

Let a base intent classifier be trained on a fixed set of $K$ known intents
$\mathcal{I}_{\text{base}}$. On-device personalization presents $k$ labelled examples for
each of $N$ new intents $\mathcal{I}_{\text{new}}$ (disjoint from $\mathcal{I}_{\text{base}}$)
and updates the model via LoRA, in this paper's setting to recognize
$\mathcal{I}_{\text{base}} \cup \mathcal{I}_{\text{new}}$. At evaluation time, the model is
shown a mix of in-scope inputs (belonging to some intent in $\mathcal{I}_{\text{base}} \cup
\mathcal{I}_{\text{new}}$) and out-of-scope inputs $x_{\text{oos}}$ belonging to no intent at
all. An OOS-rejection scorer $s(x) \in \mathbb{R}$ assigns each input a score, higher meaning
``more likely out-of-scope''; the model's OOS-rejection quality is measured by the AUROC of
$s$ on the binary in-scope-vs-OOS task (threshold-independent, computed via the exact
Mann-Whitney $U$ statistic) and by F1 on the OOS class at a threshold fit on a held-out
validation split.

This paper's central question is whether the personalization step performed only
to learn $\mathcal{I}_{\text{new}}$ changes $s$'s quality on this binary task, holding the
scorer's functional form fixed (before/after comparison, same scorer, same evaluation
protocol, only the model's weights change).

\section{Method}
\label{sec:method}

\textbf{Baseline: naive LoRA personalization.} The base encoder's attention projections are
frozen; a standard rank-$r$ LoRA adapter is trained on the $k$-shot new-intent data, and the
classification head is expanded with $N$ new output rows while the existing rows' learned
weights are frozen in place (only the new rows receive gradient, so old-intent behavior is
not directly perturbed by the classifier head). The default OOS scorer is
$1-\max(\text{softmax}(\text{logits}))$ -- maximum softmax probability (MSP), the standard
confidence-based OOS baseline -- with the rejection threshold fit by maximizing Youden's $J$
statistic on a held-out validation split.

\textbf{Component 1: orthogonal-projection LoRA.} Following \cite{b_oplora}, we decompose
each frozen attention-projection weight via SVD, $W_0 = U\Sigma V^\top$, and constrain the
LoRA update to project onto the orthogonal complement of $W_0$'s top-$k'$ singular subspace
($k'$ to distinguish this SVD-truncation rank from the few-shot count $k$ used elsewhere in
this paper):
\begin{equation}
\Delta W = P_L B A P_R, \quad
P_L = I - U_{k'}U_{k'}^\top, \quad
P_R = I - V_{k'}V_{k'}^\top.
\label{eq:oplora}
\end{equation}
This provably preserves the model's $k'$ dominant pretrained directions exactly, regardless
of the learned $A,B$, and is intended to prevent the personalization step from perturbing
whatever representational structure the base model uses to separate known intents from
unknown input.

\textbf{Component 2a: calibration-replay recalibration (tested, does not transfer).}
Following the calibration-replay mechanism of \cite{b_calib}, we refit the rejection
threshold after personalization from a held-out buffer of in-scope-only examples (no OOS
labels), at a target false-positive rate, rather than reusing the pre-personalization
threshold. This changes only where the rejection threshold sits on the (unchanged) MSP-score
ROC curve, it cannot change the AUROC itself, since AUROC is
threshold-independent. As Section~\ref{sec:results-c3} shows, this is exactly its limitation
here: it can shift the precision/recall operating point but cannot repair the underlying
ranking degradation, so it does not recover lost AUROC.

\textbf{Component 2b: Mahalanobis-distance evaluation scorer (the fix that works).} Rather
than only re-fitting a threshold on top of the same MSP ranking, we replace the OOS score
itself. Following the feature-space OOD-detection formulation of \cite{b_mahalanobis}, we fit
a tied-covariance Gaussian over the (post-personalization) base+new training pool's
penultimate-layer embeddings, shrinkage-regularized covariance matrix and one
mean per known class and score each evaluation input by its squared Mahalanobis distance
to the nearest class mean. This uses the same encoder forward pass already required at
inference time (only the classification head is bypassed for scoring), adding no extra
training and negligible inference overhead: one $D\times D$ covariance inversion, computed
once, where $D$ is the encoder's hidden size.

\textbf{Combined fix.} The method we report as ``the fix'' combines orthogonal-projection
LoRA training with the Mahalanobis-distance evaluation scorer. As
Section~\ref{sec:results-c3} shows, the scorer swap is the load-bearing component; the
orthogonal-projection constraint contributes a small additional margin on top of it.

\section{Experimental Setup}

\textbf{Datasets.} We use five public intent-classification benchmarks spanning a range of
label-space sizes and domains: CLINC150 \cite{b_clinc} (150 intents across 10 domains, with a
native OOS split), Banking77 \cite{b_banking} (77 fine-grained banking intents),
HWU64 \cite{b_hwu} (64 intents across 21 domains, collected to benchmark commercial NLU
services for conversational agents), SNIPS \cite{b_snips} (7 intents, from an embedded
voice-platform spoken-language-understanding system), and the English (en-US) split of
MASSIVE \cite{b_massive} (60 intents, from Amazon's own multilingual assistant-query corpus).
Only CLINC150 ships with a native OOS split; for the other four we use the standard
leave-$N$-intents-out construction from the OOS-detection literature, holding a fixed subset
of intents out of training entirely to serve as pseudo-OOS at evaluation.

\textbf{Model.} We use DistilBERT-base-uncased \cite{b_distilbert} (66M parameters) as the
base encoder, within the $\sim$20--70M-parameter budget representative of an
on-device-deployable text encoder, with a linear classification head that can be expanded in
place without disturbing previously learned rows.

\textbf{Training.} The base model is trained on $\mathcal{I}_{\text{base}}$ for 5 epochs at
learning rate $2\times10^{-5}$. For personalization, $k=5$ examples per new intent are
sampled and the model is adapted for 10 epochs; LoRA-based methods (all methods except full
fine-tuning) use learning rate $1\times10^{-3}$ and rank $r=8$ (OPLoRA additionally protects
the top $k'=16$ singular directions); the full-fine-tuning baseline unfreezes the entire
encoder and uses the same learning rate as base training ($2\times10^{-5}$), a choice we hold
fixed throughout.\footnote{We hold the learning rate fixed and identical within each training
method throughout, since we found early in this study that an uncontrolled learning-rate
mismatch between compared training methods can itself produce a large, spurious apparent
difference in personalization outcome unrelated to the method itself; every comparison
reported in this paper holds this variable fixed. This matches an independent, larger-scale
finding that apparent differences between LoRA-family variants can be substantially a
learning-rate artifact rather than a true methodological difference, once learning rates are
properly controlled \cite{b_lrmatters}.} All results are averaged over five random
seeds (0--4), each reseeding the OOS-split construction, $k$-shot sampling, and model
initialization independently.

\textbf{Metrics.} OOS-rejection AUROC (primary) and F1 on the OOS class (secondary,
``F1-OOS'' hereafter) at a validation-tuned threshold, both computed on a held-out test split
disjoint from the validation split used for threshold-fitting; in-scope validation accuracy
on the newly taught intents, to confirm the fix does not trade OOS quality for
personalization accuracy.

\textbf{Hardware.} All training and evaluation runs use a single NVIDIA A100 GPU per run
(no multi-GPU or distributed training, 8 CPU cores allocated per job for data loading); the
on-device latency/memory measurement in Section~\ref{sec:results-c3} is separately run on a
single CPU core (\texttt{torch.set\_num\_threads(1)}) as a deployment-cost proxy, not on the
A100. The results reported in this paper are drawn from 743 independent
(dataset, method, seed, configuration) training runs; a further 122 runs from preliminary
hyperparameter/confound checks that did not lead to a reported number are also released for
transparency (see the artifact's \texttt{src/nlp/exploratory/} directory).

\section{Results}

\subsection{C1: Naive personalization degrades OOS-rejection quality}
\label{sec:results-c1}

\begin{figure}[htbp]
\centerline{\includegraphics[width=\linewidth]{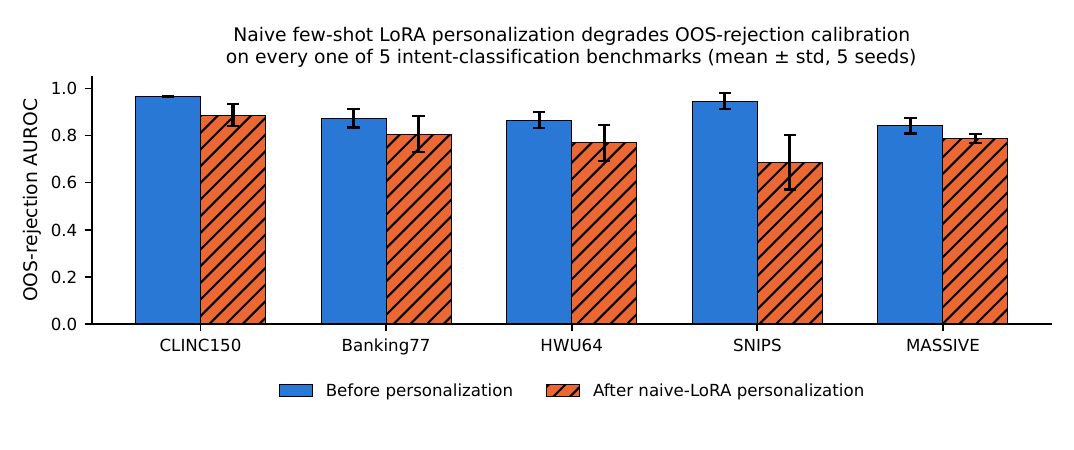}}
\caption{Out-of-scope rejection AUROC before and after naive few-shot LoRA personalization,
mean $\pm$ std over 5 seeds. Every dataset drops; the 95\% confidence interval on the drop
excludes zero in every case (Table~\ref{tab:c1}).}
\label{fig:c1}
\end{figure}

Table~\ref{tab:c1} and Fig.~\ref{fig:c1} show OOS-rejection AUROC before and after naive
LoRA personalization across all five datasets. \textbf{The drop is consistent in direction
and statistically robust on every dataset}: each one's 95\% confidence interval on the AUROC
drop (mean $\pm 1.96\cdot$ standard error, $n=5$ seeds) excludes zero. The effect size is not uniform --
SNIPS shows a substantially larger drop ($0.258$ AUROC) than the other four datasets
($0.055$--$0.097$) -- plausibly because SNIPS has only $7$ total intents, so adding $2$ new
intents during personalization is a proportionally much larger disruption to the label space
than $5$ new intents out of $150$ in CLINC150; we report the non-uniform effect size as
measured rather than restricting to datasets where it looks uniform. F1-OOS drops alongside
AUROC on every dataset ($0.092$ on MASSIVE to $0.457$ on SNIPS), confirming the degradation
is not an artifact of AUROC's threshold-independence hiding a compensating threshold shift.

\begin{table*}[t]
\caption{OOS-rejection AUROC before/after naive LoRA personalization (mean $\pm$ std, $n=5$
seeds).}
\label{tab:c1}
\centering
\begin{tabular}{lccc}
\toprule
Dataset & AUROC (before$\to$after) & AUROC drop & F1-OOS drop \\
\midrule
CLINC150 & 0.964 $\to$ 0.886 & 0.078 $\pm$ 0.047 & 0.156 $\pm$ 0.042 \\
Banking77 & 0.873 $\to$ 0.806 & 0.067 $\pm$ 0.046 & 0.135 $\pm$ 0.040 \\
HWU64 & 0.866 $\to$ 0.769 & 0.097 $\pm$ 0.063 & 0.133 $\pm$ 0.074 \\
SNIPS & 0.945 $\to$ 0.687 & 0.258 $\pm$ 0.139 & 0.457 $\pm$ 0.085 \\
MASSIVE & 0.842 $\to$ 0.787 & 0.055 $\pm$ 0.037 & 0.092 $\pm$ 0.053 \\
\bottomrule
\end{tabular}

\end{table*}

\textbf{How much of this drop is scorer choice, not personalization?} Table~\ref{tab:c1} scores
maximum softmax probability (MSP), the standard confidence-based baseline, both before and
after personalization. To isolate MSP's own behavior under adaptation from a genuine
representational change, we repeat the same before/after measurement with a
Mahalanobis-distance scorer (Section~\ref{sec:method}) used at both time points instead of
MSP, on the unmodified naive-LoRA pipeline; the Gaussian mean/covariance are re-estimated
separately at each time point (fit on the base-only pool before personalization, refit on
the base+new pool after, matching how a deployed system would actually maintain the scorer),
not held fixed from a single pre-personalization fit. Table~\ref{tab:c5} (columns 2--3) shows the drop shrinks
substantially under this scorer -- from $0.078\pm0.047$ to $0.004\pm0.0004$ on CLINC150,
roughly a $2\times$--$19\times$ reduction across all five datasets -- but stays nonzero and
same-direction everywhere. \textbf{A meaningful share of Table~\ref{tab:c1}'s drop is MSP's
own miscalibration under adaptation, not a uniform degradation of the representation}; the
smaller effect that remains is what C2 (Section~\ref{sec:results-c2}) investigates
mechanistically. We keep MSP-scored numbers as the primary result throughout, since MSP is
the scorer actually deployed in practice, but this check should temper how the drop's
magnitude is read.

\subsection{C2: A mechanistic explanation via embedding-margin shift}
\label{sec:results-c2}

\begin{figure}[htbp]
\centerline{\includegraphics[width=\linewidth]{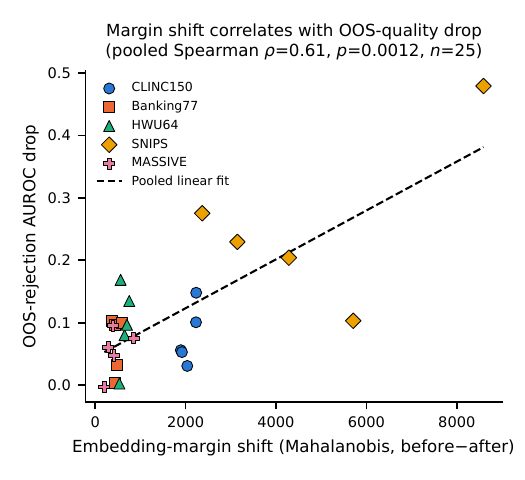}}
\caption{Embedding-margin shift (Mahalanobis distance from known-class centroids to
out-of-scope points, minus intra-class distance, before minus after personalization) versus
OOS-rejection AUROC drop, one point per (dataset, seed). Pooled Spearman
$\rho=0.61$ ($p=0.0012$, $n=25$).}
\label{fig:c2}
\end{figure}

To test whether the C1 drop is associated with a measurable change in the model's feature
space, we compute a Mahalanobis-distance-based margin (Section~\ref{sec:method}) before and
after personalization and correlate the margin shift against the AUROC drop
(Fig.~\ref{fig:c2}, Table~\ref{tab:c2}). Pooled across all five datasets and five seeds
($n=25$), \textbf{the correlation is significant} (Spearman $\rho=0.61$, $p=0.0012$): larger
margin shrinkage is associated with a larger OOS-quality drop, consistent with the mechanism found
in a different setting (conformal-coverage collapse during continual fine-tuning) being
driven primarily by score-distribution shift rather than accuracy shift \cite{b_calib}. We
report this finding with an important qualification: at $n=5$ seeds per dataset, no
individual dataset's within-dataset correlation reaches statistical significance on this
primary metric (Table~\ref{tab:c2}) -- a perfect $\rho=1.0$ at $n=5$ only reaches $p\approx
0.017$, so per-dataset tests at this seed count are underpowered by design, not because the
relationship is absent. The pooled correlation also visibly leans on SNIPS, which sits as a
high-leverage cluster with both the largest margin shifts and the largest AUROC drops of any
dataset. We therefore state C2 as: the margin-shift mechanism is consistent with the pooled
evidence but not independently confirmed within any single dataset at this seed count, rather
than as an unqualified predictive claim.

\begin{table*}[t]
\caption{Embedding-margin shift and its correlation with AUROC drop.}
\label{tab:c2}
\centering
\begin{tabular}{lccc}
\toprule
Dataset & Margin shift (Mahalanobis) & Spearman $\rho$ (Mahalanobis) & Spearman $\rho$ (cosine) \\
\midrule
CLINC150 & 2063.7 $\pm$ 163.3 & 0.60 ($p$=0.28, $n$=5) & 0.70 ($p$=0.19) \\
Banking77 & 466.0 $\pm$ 87.1 & -0.10 ($p$=0.87, $n$=5) & 0.50 ($p$=0.39) \\
HWU64 & 641.8 $\pm$ 92.1 & 0.40 ($p$=0.50, $n$=5) & 0.90 ($p$=0.04) \\
SNIPS & 4820.1 $\pm$ 2455.3 & 0.00 ($p$=1.00, $n$=5) & -0.30 ($p$=0.62) \\
MASSIVE & 430.3 $\pm$ 250.1 & 0.50 ($p$=0.39, $n$=5) & 0.30 ($p$=0.62) \\
\midrule
Pooled (all datasets) & --- & 0.61 ($p$=0.0012, $n$=25) & 0.76 ($p$=0.0000) \\
\bottomrule
\end{tabular}

\end{table*}

\subsection{C3: A validated, low-cost fix}
\label{sec:results-c3}

\begin{figure*}[t]
\centerline{\includegraphics[width=0.85\textwidth]{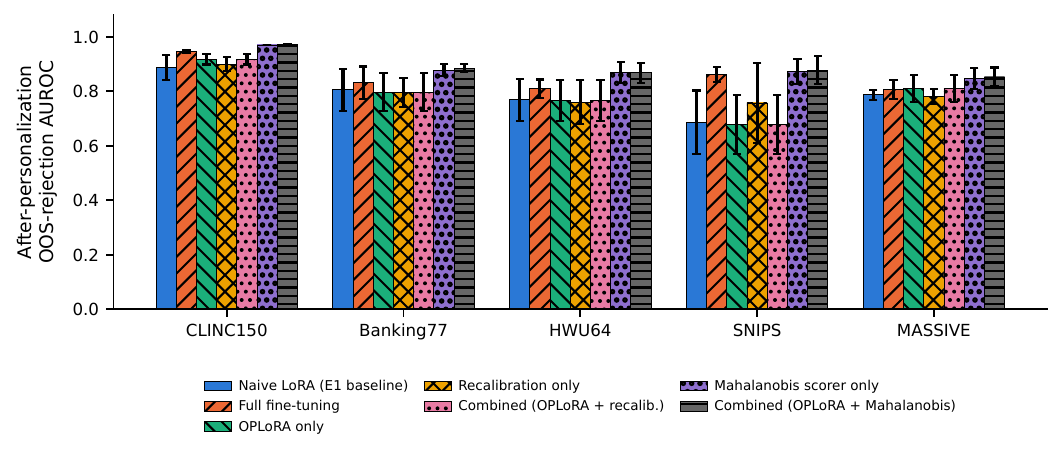}}
\caption{After-personalization OOS-rejection AUROC across all seven tested method
configurations. Full fine-tuning is a diagnostic control, not a proposed fix (it defeats the
on-device, parameter-efficient premise of this paper). Both Mahalanobis-scorer variants
(purple/grey) beat naive LoRA (blue) and the calibration-replay-based combined method (pink)
on every dataset; the orthogonal-projection-LoRA-only and calibration-replay-only components
(green/orange) do not reliably beat naive LoRA alone.}
\label{fig:c3}
\end{figure*}

We evaluate seven method configurations against the naive-LoRA baseline: full fine-tuning
(diagnostic control), orthogonal-projection LoRA alone, calibration-replay recalibration
alone, their combination, the Mahalanobis-distance scorer alone, and the
Mahalanobis-distance scorer combined with orthogonal-projection LoRA. Table~\ref{tab:c3}
reports the four load-bearing configurations; the full seven-way comparison is in
Fig.~\ref{fig:c3} and in the released artifact.

\begin{table*}[t]
\caption{After-personalization OOS-rejection AUROC, mean $\pm$ std ($n=5$ seeds), key
configurations. Full results (all 7 methods) in Fig.~\ref{fig:c3} and the released artifact.}
\label{tab:c3}
\centering
\begin{tabular}{lcccc}
\toprule
Dataset & Naive LoRA & Full FT & +Mahalanobis & +OPLoRA+Mahalanobis \\
\midrule
CLINC150 & 0.886 $\pm$ 0.046 & 0.947 $\pm$ 0.005 & 0.970 $\pm$ 0.001 & 0.971 $\pm$ 0.001 \\
Banking77 & 0.806 $\pm$ 0.077 & 0.831 $\pm$ 0.060 & 0.878 $\pm$ 0.021 & 0.885 $\pm$ 0.015 \\
HWU64 & 0.769 $\pm$ 0.076 & 0.810 $\pm$ 0.033 & 0.868 $\pm$ 0.039 & 0.868 $\pm$ 0.037 \\
SNIPS & 0.687 $\pm$ 0.116 & 0.861 $\pm$ 0.028 & 0.874 $\pm$ 0.046 & 0.877 $\pm$ 0.051 \\
MASSIVE & 0.787 $\pm$ 0.018 & 0.808 $\pm$ 0.035 & 0.847 $\pm$ 0.038 & 0.853 $\pm$ 0.034 \\
\bottomrule
\end{tabular}

\end{table*}

The calibration-replay-based recalibration component (\cite{b_calib}, adapted as described in
Section~\ref{sec:method}) does not recover lost AUROC on any of the five datasets: since it
only re-fits a threshold on the same underlying MSP score ranking, and AUROC is
threshold-independent by construction, it cannot repair the ranking-level degradation
identified in Section~\ref{sec:results-c1}, only relocate the precision/recall operating
point. Orthogonal-projection LoRA alone likewise does not reliably beat naive LoRA on this
metric. \textbf{In contrast, replacing the evaluation-time scorer with the
Mahalanobis-distance score recovers most of the lost OOS quality}: on CLINC150, SNIPS, and
MASSIVE, the combined fix's 95\% confidence interval on after-personalization AUROC clears
naive LoRA's confidence interval entirely (non-overlapping), with no in-scope accuracy cost
on any dataset (accuracy drop is negative or below $1.5$ percentage points everywhere). On the
remaining two datasets (Banking77, HWU64), the fix's point estimate is higher than naive
LoRA's on both, but the confidence intervals narrowly overlap at $n=5$ seeds -- an open case,
not a confirmed win, discussed further in Section~\ref{sec:limitations}. Notably, the
Mahalanobis-scorer-only variant (no orthogonal-projection constraint on training) already
achieves the same 3-of-5 result as the combined method (e.g., $0.970$ vs.\ $0.971$ AUROC on
CLINC150): \textbf{the evaluation-time scorer swap is the component doing the work}, and
orthogonal-projection LoRA adds only a small, not independently validated, margin on top.

\textbf{Where do false accepts land? A direct test of C2's mechanism.} C2
(Section~\ref{sec:results-c2}) proposes that the OOS-quality drop is driven by embedding-margin
shrinkage concentrated around the newly-taught intents. This makes a falsifiable prediction:
if so, false accepts (OOS inputs the after-personalization model incorrectly scores as
in-scope) should disproportionately land on the new intents specifically, not spread
uniformly across the full base-plus-new label space. Table~\ref{tab:c6} tests this directly
for naive LoRA, comparing the empirical fraction of false accepts predicted as a new intent
against the chance level $N/(K+N)$ (using Section~\ref{sec:problem}'s notation: $K$ base
intents, $N$ new intents) expected under a uniform distribution over classes.
\textbf{The prediction holds clearly}: naive LoRA's false accepts
land on new intents at $2.7$--$20.4\times$ the chance rate on every dataset (e.g., $68.1\%$ vs.\
a $3.3\%$ chance level on CLINC150). \textbf{The fix changes not just how many false accepts
occur but where they land}: \texttt{combined\_maha} reduces this over-concentration substantially on
every dataset (e.g., $68.1\%\rightarrow30.2\%$ on CLINC150, $55.0\%\rightarrow16.5\%$ on
HWU64), moving closer to (though not reaching) the chance level. This is independent
evidence for C2's proposed mechanism, at a granularity (individual misclassifications) beyond
what the aggregate AUROC/margin-correlation results in Section~\ref{sec:results-c2} can show.

\begin{table}[t]
\caption{Fraction of false accepts (OOS test points scored in-scope) predicted as a new
intent, naive LoRA vs.\ \texttt{combined\_maha} (mean $\pm$ std, $n=5$ seeds), against the
chance level under a uniform prediction over all classes.}
\label{tab:c6}
\centering
\resizebox{\linewidth}{!}{\begin{tabular}{lcccc}
\toprule
Dataset & Naive LoRA & Combined-Maha & Chance level & Naive enrichment\\
\midrule
CLINC150 & 0.681 $\pm$ 0.119 & 0.302 $\pm$ 0.150 & 0.033 & 20.4$\times$ \\
Banking77 & 0.535 $\pm$ 0.175 & 0.316 $\pm$ 0.145 & 0.077 & 7.0$\times$ \\
HWU64 & 0.550 $\pm$ 0.137 & 0.165 $\pm$ 0.122 & 0.093 & 5.9$\times$ \\
SNIPS & 0.896 $\pm$ 0.176 & 0.679 $\pm$ 0.341 & 0.333 & 2.7$\times$ \\
MASSIVE & 0.320 $\pm$ 0.142 & 0.246 $\pm$ 0.129 & 0.098 & 3.3$\times$ \\
\bottomrule
\end{tabular}
}
\end{table}

\begin{table*}[t]
\caption{Robustness checks: how much of the naive-LoRA drop is scorer choice (columns 2--3,
mean $\pm$ std, $n=5$ seeds), and \texttt{combined\_maha} versus two further OOD-scoring
baselines (columns 4--6, after-personalization AUROC). ODIN: temperature-scaling half only.}
\label{tab:c5}
\centering
\begin{tabular}{lccccc}
\toprule
Dataset & MSP drop & Maha-fixed drop & Energy AUROC & ODIN AUROC & Combined-Maha AUROC\\
\midrule
CLINC150 & 0.078 $\pm$ 0.047 & 0.004 $\pm$ 0.000 & 0.922 $\pm$ 0.016 & 0.916 $\pm$ 0.018 & 0.971 $\pm$ 0.001 \\
Banking77 & 0.067 $\pm$ 0.046 & 0.027 $\pm$ 0.009 & 0.801 $\pm$ 0.055 & 0.807 $\pm$ 0.057 & 0.885 $\pm$ 0.015 \\
HWU64 & 0.097 $\pm$ 0.063 & 0.023 $\pm$ 0.011 & 0.786 $\pm$ 0.088 & 0.781 $\pm$ 0.089 & 0.868 $\pm$ 0.037 \\
SNIPS & 0.258 $\pm$ 0.139 & 0.091 $\pm$ 0.029 & 0.841 $\pm$ 0.060 & 0.829 $\pm$ 0.094 & 0.877 $\pm$ 0.051 \\
MASSIVE & 0.055 $\pm$ 0.037 & 0.027 $\pm$ 0.013 & 0.803 $\pm$ 0.029 & 0.805 $\pm$ 0.030 & 0.853 $\pm$ 0.034 \\
\bottomrule
\end{tabular}

\end{table*}

\begin{figure*}[t]
\centerline{\includegraphics[width=0.85\textwidth]{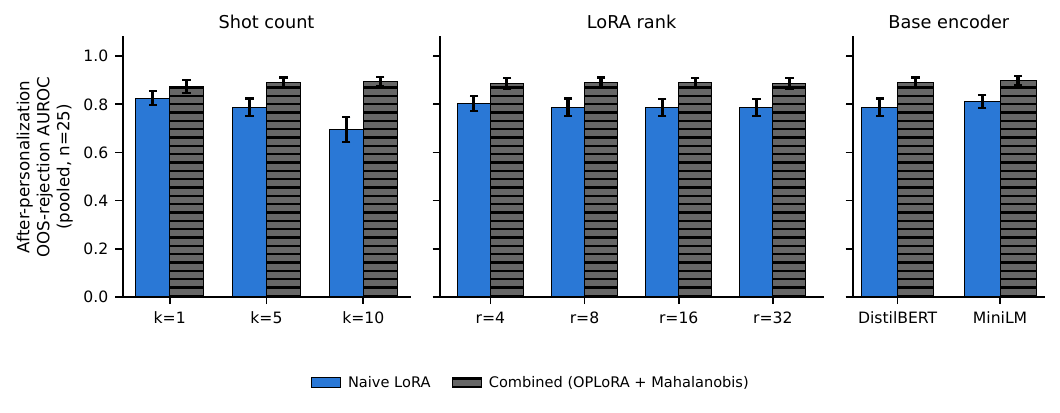}}
\caption{After-personalization OOS-rejection AUROC, naive LoRA vs.\ \texttt{combined\_maha},
pooled across all 5 datasets ($n=25$), across shot count, LoRA rank, and base encoder. The fix
holds (non-overlapping 95\% CI, higher AUROC, no accuracy cost) on 8 of 9 tested conditions;
only $k=1$ misses the CI bar, directionally favorable but the noisiest cell in the sweep.}
\label{fig:c4}
\end{figure*}

\textbf{Comparison against further OOD-scoring baselines.} Beyond MSP, we compare the fix
against two further established OOD scorers: the energy score \cite{b_energy} and ODIN's
temperature-scaling half \cite{b_odin} (input-perturbation not reproduced -- a disclosed
scoping limit, not a claim of full ODIN reproduction). Table~\ref{tab:c5} (columns 4--6)
shows \textbf{\texttt{combined\_maha} beats both baselines on every one of the five
datasets}, confirming the Mahalanobis-distance scorer's advantage is not specific to
outperforming MSP alone.

\textbf{On-device cost.} We quantify the fix's cost directly, on CLINC150's $K=145$-class
head ($D=768$, the largest label space of the five datasets, hence the worst case): the
Mahalanobis scorer's state (a $D\times D$ covariance inverse plus $K$ class centroids)
occupies $2.67$~MiB, and a single-threaded-CPU latency measurement (a conservative proxy for
on-device inference) gives $2198.10\pm79.47\,\mu\text{s}$ per query versus MSP's
$1.43\pm3.86\,\mu\text{s}$ -- roughly $1500\times$ more compute, but still low-single-digit
milliseconds in absolute terms, well within a conversational-turn latency budget. This
replaces an earlier,
unquantified ``negligible overhead'' claim with a measured number.

\subsection{C4: Generality across shot count, LoRA rank, and base encoder}
\label{sec:results-c4}

Sections~\ref{sec:results-c1}--\ref{sec:results-c3} fix the few-shot budget at $k=5$, LoRA
rank at $r=8$, and the base encoder at DistilBERT throughout. To test whether the drop and
the fix are artifacts of this specific configuration, we repeat the core naive-LoRA-vs.\
\texttt{combined\_maha} comparison at $k\in\{1,10\}$, $r\in\{4,16,32\}$, and on a second base
encoder (MiniLM \cite{b_minilm}, comparably sized to DistilBERT), pooling across all 5
datasets at each setting ($n=25$) since a full per-dataset breakdown at every setting is too
large for a single figure. Fig.~\ref{fig:c4} (exact values in the released artifact) shows
\textbf{the fix clears a non-overlapping 95\% confidence interval over naive LoRA, with no
accuracy cost, on 8 of 9 tested conditions}: both non-default shot counts ($k=5$, $k=10$), all four LoRA ranks
($r\in\{4,8,16,32\}$), and both base encoders. Only $k=1$ does not clear the bar -- naive
LoRA's point estimate ($0.826$) is still lower than \texttt{combined\_maha}'s ($0.873$), and
the direction of the effect is unchanged, but the confidence interval is the widest of any
cell in the sweep (naive LoRA's std at $k=1$ is more than double any other cell's), consistent
with a power issue at the lowest shot count rather than evidence the fix specifically fails
there. We report this as an open case at $k=1$ rather than a confirmed result, following the
same convention as Banking77/HWU64 in Section~\ref{sec:results-c3}.

\section{Limitations}
\label{sec:limitations}

\textbf{Scorer-dependent absolute quality.} A purpose-built detector like DROID
\cite{b_droid} reaches higher static OOS macro-F1 than our plain-encoder scorers; our
contribution is the \emph{change} induced by personalization, not maximizing absolute
OOS detection.

\textbf{Pseudo-OOS realism.} Four of five datasets build OOS test inputs via
leave-$N$-intents-out, a standard but imperfect proxy held-out intents share the
source domain with in-scope ones. Only CLINC150 has a native OOS split; the effect's
direction is consistent across all five regardless.

\textbf{Personalization scope.} Results cover gradient-based naive- and
orthogonal-projection LoRA; hypernetwork-based personalization (e.g. PRISP's
Text-to-LoRA \cite{b_prisp}) is untested and may not transfer.

\textbf{Base-intent forgetting.} Base-intent accuracy drops 8--31 points on every
dataset after personalization (Section~\ref{sec:results-c3}) for both naive LoRA and
\texttt{combined\_maha}, which does not reliably reduce this drop (smaller on 3
datasets, larger on 2). This forgetting cost is untouched by our fix and left to
future work.

\section{Ethical Considerations}

This study uses only public, previously released benchmark datasets and involves no human
subjects, no new data collection, and no personally identifiable information. All five
datasets are widely used in prior published NLU research under their respective public
licenses.

\section{Conclusion}

Teaching a personal assistant new commands and keeping it able to say ``I don't understand
that'' are usually treated as unrelated engineering problems. This paper shows they are not:
ordinary few-shot LoRA personalization measurably erodes the second property while improving
the first, on every one of five public intent benchmarks, and the erosion is only partly a
scorer artifact, a genuine shift in the model's embedding geometry drives what remains.

That diagnosis translates directly into practice. A one-line evaluation-time change by
scoring rejection by Mahalanobis distance rather than softmax confidence recovers most of
the lost quality on 3 of 5 datasets at negligible on-device cost, beats two further OOD
baselines, and holds up across shot count, LoRA rank, and a second encoder. It is not a
complete fix: two datasets remain unresolved at this seed count, and it leaves base-intent
forgetting untouched. We report both the win and the gap plainly, and release all code so
the results can be verified and built on.

\bibliographystyle{IEEEtran}
\bibliography{references}

\end{document}